\documentclass[10pt,twocolumn,letterpaper]{article}

\usepackage{wacv}              

\usepackage[dvips]{graphicx}
\usepackage{amsmath}
\usepackage{amssymb}
\usepackage{amsfonts}
\usepackage{booktabs}
\usepackage{float}
\usepackage{multirow}
\usepackage{pifont}
\usepackage[dvipsnames]{xcolor}
\usepackage{dcolumn}
\usepackage{caption}
\usepackage{enumitem}
\usepackage{tabularx}
\usepackage{afterpage}
\usepackage{hhline}
\usepackage[accsupp]{axessibility}
\usepackage[toc,page]{appendix}

\usepackage[pagebackref,breaklinks,colorlinks]{hyperref}

\usepackage[capitalize]{cleveref}
\crefname{section}{Sec.}{Secs.}
\Crefname{section}{Section}{Sections}
\Crefname{table}{Table}{Tables}
\crefname{table}{Tab.}{Tabs.}

\newcommand{\codename}[0]{{FastSR-NeRF}}

\def\wacvPaperID{2322} 
\def\confName{WACV}
\def\confYear{2024}

\makeatletter
\apptocmd\@maketitle{{\teaser{}\par}}{}{}
\makeatother

\begin{document}


\title{\codename{}: Improving NeRF Efficiency on Consumer Devices with \\ A Simple Super-Resolution Pipeline}

\author{Chien-Yu Lin\thanks{Work done while interning at Apple.}\\
    University of Washington\\
    Seattle, WA, USA\\
    {\tt\small cyulin@cs.washington.edu}
\and
Qichen Fu,
Thomas Merth,
Karren Yang,
Anurag Ranjan\\
Apple, Inc.\\
Cupertino, CA, USA\\
{\tt\small \{qfu22,tmerth,karren\_yang,anuragr\}@apple.com}
}

\newcommand\teaser{%
    \centering
    \includegraphics[width=0.9\textwidth]{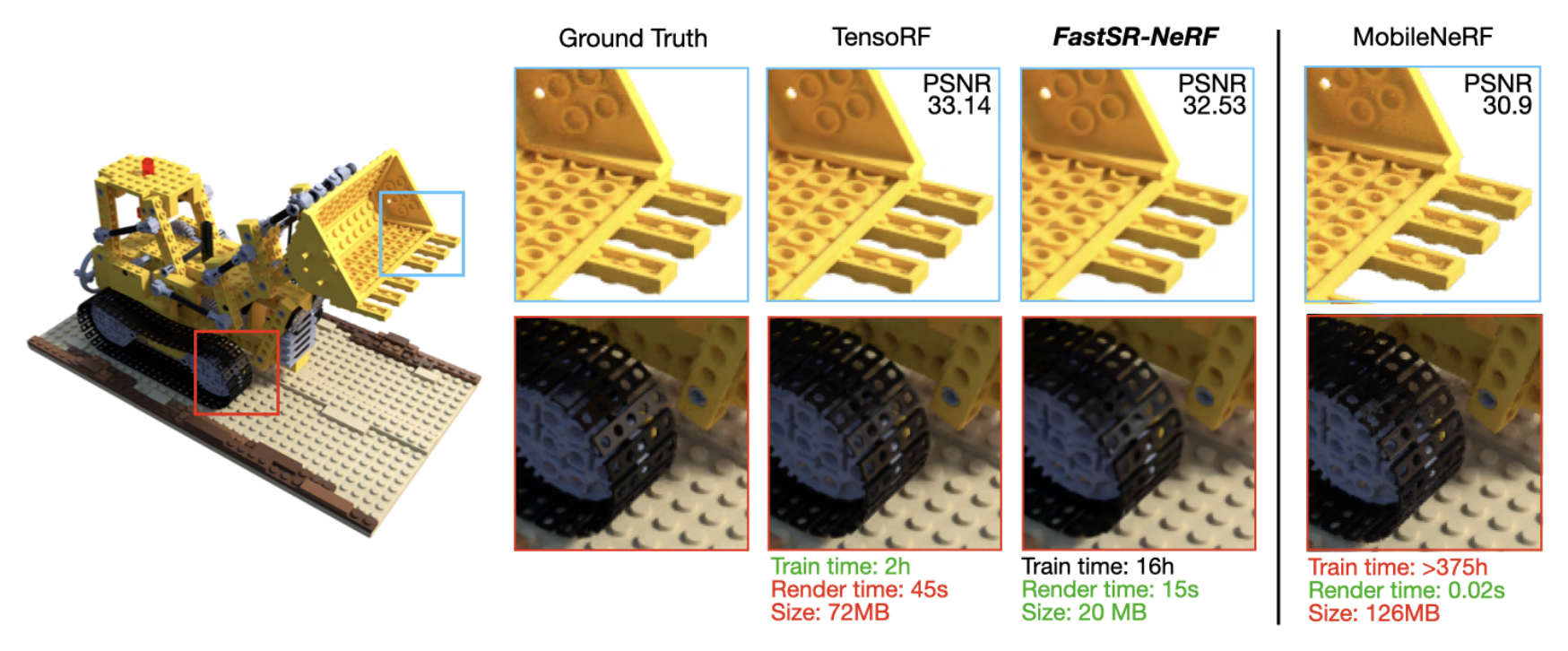}
    \captionof{figure}{\textbf{Super-resolution (SR) can be used to improve neural rendering efficiency under a limited training budget.} Comparison of TensoRF, \codename{} (ours), and MobileNeRF \cite{chen2022mobilenerf} on a consumer-grade MacBook Air M2 laptop. 
    \codename{} employs a straightforward SR pipeline (TensoRF+SR), which can enhance rendering times and compress the model size while incurring relatively low training overhead. While state-of-the-art mobile models such as MobileNeRF can render very quickly, they cannot be trained on consumer devices under a meaningful time budget.
    }
    \label{fig:teaser}
    \vspace{20pt}
}
\maketitle


\begin{abstract}
    Super-resolution (SR) techniques have recently been proposed to upscale the outputs of neural radiance fields (NeRF) and generate high-quality images with enhanced inference speeds. 
    However, existing NeRF+SR methods increase training overhead by using extra input features, loss functions, and/or expensive training procedures such as knowledge distillation. 
    In this paper, we aim to leverage SR for efficiency gains without costly training or architectural changes.
    Specifically, we build a simple NeRF+SR pipeline that directly combines existing modules, and we propose a lightweight augmentation technique, random patch sampling, for training.
    Compared to existing NeRF+SR methods, our pipeline mitigates the SR computing overhead and can be trained up to 23$\times$ faster, making it feasible to run on consumer devices such as the Apple MacBook.
    Experiments show our pipeline can upscale NeRF outputs by 2-4$\times$ while maintaining high quality, increasing inference speeds by up to 18$\times$ on an NVIDIA V100 GPU and 12.8$\times$ on an M1 Pro chip.
    We conclude that SR can be a simple but effective technique for improving the efficiency of NeRF models for consumer devices. 
    \vspace{-15pt}
\end{abstract}

\section{Introduction}
\label{sec:intro}

Neural Radiance Field (NeRF) models \cite{mildenhall2020nerf} have become integral to many computer vision and computer graphics tasks, such as novel view synthesis \cite{mildenhall2020nerf, mipnerf, liu2020nsvf}, surface reconstruction \cite{wang2021neus, yariv2021volume}, camera pose estimation \cite{YenChen20arxiv_iNeRF, Wang21arxiv_nerfmm} and 3D image generation \cite{Chan2021_eg3d, poole2022dreamfusion, lin2023magic3d}. 
Since the original NeRF model cannot render images efficiently, a large body of research \cite{killonerf, liu2020nsvf, fridovich2022plenoxels, chen2022tensorf, sun2021dvgo, muller2022instant, Wu_2023_CVPR_nffb} has been dedicated to address the rendering efficiency. 
Many of these works achieve impressive gains by decomposing and representing the 3D neural radiance field with explicit features \cite{hedman2021snerg, fridovich2022plenoxels, liu2020nsvf, chen2022mobilenerf, sun2021dvgo, chen2022tensorf, muller2022instant}. 
However, these methods often require extended training times and/or specialized architectures and kernel support on high-end GPUs. 
For example, MobileNeRF is capable of fast rendering on mobile devices \cite{chen2022mobilenerf}, but uses 8 server-class GPUs for training \cite{mobilenerf_code}, which translates to over 15 days ($>$375h) on a consumer-grade laptop (Figure \ref{fig:teaser}, right).
To improve the accessibility and personalized use of NeRFs, there is a need to explore efficient rendering techniques that can also be trained on consumer-grade devices.


In this paper, we introduce \codename{}, which demonstrates CNN-based super-resolution (SR) can be a simple, low-cost technique for improving the efficiency of NeRF models on consumer devices. 
The basic idea 
is to train a small NeRF model to generate lower-resolution scene features with 3D consistency, and a fast SR model to generate higher-resolution features. 
%
This combination reduces the number of pixels that need to be computed using the NeRF's slow volume rendering process, increasing rendering speed.
%
While SR techniques for neural rendering have been proposed by previous works, these methods either (i) involve specialized SR modules that use extra input features such as high-resolution reference images \cite{huang2023refsrnerf}; (ii) employ expensive training procedures such as distillation \cite{cao2022mobiler2l}; or (iii) are trained on tens of thousands of images within a generative modeling framework \cite{Chan2021_eg3d}. 
%
None of these methods can be feasibly trained on low-power, consumer-grade platforms. Whether it is possible to achieve high-quality results with SR under a limited training budget remains an open question.

Here, we address this question by exploring a simple NeRF+SR pipeline that directly combines existing modules. We hypothesize that the spatial inductive bias of CNN-based SR is sufficient to generate high-quality outputs for low upscaling ratios, even without extra input features or complex training procedures.
%
%
%
%
To improve synthesis quality, we propose only a lightweight augmentation technique called \textbf{\textit{random patch sampling}}:
rather than extract patches from an image grid for training the SR module as done in existing works\cite{huang2023refsrnerf,wang20224k-nerf}, we extract patches from random positions to increase the diversity of image patches seen by the SR module. 
%
Experiments across three datasets show, somewhat surprisingly, our simple
NeRF+SR pipeline with low training overhead can achieve
comparable quality and greater rendering efficiency than existing complex NeRF+SR pipelines.
To summarize, the key results of our study are as follows:

\begin{itemize}[leftmargin=*]
\item \textbf{SR can be a nearly ``free'' technique for improving neural rendering efficiency.} Our comprehensive experiments across three datasets show that applying SR to a NeRF model at 2-4× upscaling ratios, without any complex training procedures or architectural modifications, can improve inference speeds by up to 35.7× on an NVIDIA V100 GPU and 12.8× on an M1 Pro chip, while maintaining peak signal-to-noise ratio (PSNR) in a 0.4-1.2 dB range. \emph{Surprisingly, our simple pipeline can achieve comparable quality to recent and more complex SR techniques \cite{wang2021nerf-sr, huang2023refsrnerf}, while being more efficient in training and inference.}

\item \textbf{Random patch sampling is a crucial lightweight augmentation technique for NeRF+SR.} We propose random patch sampling, a lightweight augmentation technique. This augmentation improves the PSNR of the SR module by up to 0.89 dB for 2× upscaling and up to 1.44 dB for 4× upscaling compared to standard grid-based patch sampling, outperforming expensive distillation approaches \cite{cao2022mobiler2l} at a fraction of the time cost.

\item \textbf{\codename{} is one of the few efficient methods that can be \emph{trained} on a low-power device.} 
As shown in Figure \ref{fig:teaser}, by utilizing a simple NeRF+SR pipeline, \codename{} can be trained on consumer devices such as a MacBook Air M2, whereas most other models and existing NeRF+SR pipelines fail to train with a meaningful time budget. 

\end{itemize}

Overall, our analysis shows that SR can be a low-cost, plug-and-play strategy for improving the efficiency of neural rendering models under a limited training budget. Even a simple NeRF+SR pipeline can make neural rendering more efficient and accessible for those with low-power, consumer-grade hardware.

\section{Background}
\label{sec:background}


\subsection{Neural Radiance Fields}
The NeRF model was first proposed in \cite{mildenhall2020nerf}. 
Given a position and view angle in a 3D scene, NeRF uses a large MLP network to map from the 5D input (3D coordinates plus 2D view angle) to an RGB and a density value. To render a 2D image, these MLP outputs are integrated along rays passing through each pixel using volume rendering. The MLP is optimized using gradient descent with respect to a photometric loss over a sparse set of scene-specific images. 
Due to its impressive results on static novel view synthesis, NeRF quickly propelled the state-of-the-art in many other fields, including 3D image generation \cite{poole2022dreamfusion, lin2023magic3d, Chan2021_eg3d}, 3D scene editing \cite{instructnerf2023} and landscape reconstruction \cite{xiangli2022bungeenerf}. 
However, the drawback of the original NeRF is that it takes a long time to render images due to the slow volume rendering process.

To address this issue, many works have since been proposed to improve NeRF's rendering efficiency.
One line of work \cite{garbin2021fastnerf, yu2021plenoctrees, hedman2021snerg, chen2022mobilenerf} maps the learned radiance field to explicit representations such as octree-based \cite{yu2021plenoctrees} or voxel-based \cite{hedman2021snerg} data structures.
These methods achieve faster rendering tine at the cost of larger memory and training time requirements.
Another line of work \cite{Hu_2022_CVPR_efficient_nerf, fang2021neusample, li2023nerfacc, muller2022instant} focuses on improving the sampling algorithm to reduce overall computation, which yields a modest amount of acceleration.
An emerging series of works divides the radiance field into explicit voxels \cite{liu2020nsvf, killonerf, sun2021dvgo}, or some efficient representation such as matrix decomposition \cite{chen2022tensorf}, hash table \cite{muller2022instant}, and tri-plane \cite{Chan2021_eg3d, kplanes_2023, Cao2023HexPlane}.
As these models usually make use of a mix of explicit representations and MLP, they are referred as hybrid NeRFs.
Typically, hybrid NeRF models can highly accelerate training time and rendering speed, but need a relatively large model size.
Among these efficient NeRF methods, there is a trend to develop customized GPU kernels to further accelerate the specialized operations designed for each method.
Although using customized kernels can bring a major speedup, it limits the ability to deploy the model on different classes of GPUs and consumer-grade devices \cite{cao2022mobiler2l}.
Orthogonal to these works, we explore the application of SR modules for improving NeRF rendering efficiency under a limited training budget. To maximize flexibility for running on low-end devices, we consider Python implementations that do not use customized GPU kernels.


\subsection{Super Resolution with NeRFs}
Super-resolution (SR) is a recent, still under-explored method for enhancing NeRF efficiency.
EG3D \cite{Chan2021_eg3d} applies SR on top of volume rendering within a generative adversarial network for 3D faces. The SR module in their network is trained on tens of thousands of images, whereas we consider scene-specific optimization on much smaller datasets (\emph{e.g.,} 20-200 images per scene). 
NeRF-SR \cite{wang2021nerf-sr} performs sub-pixel sampling to super-resolve outputs, but this requires more compute and thus a longer training time.
MobileR2L \cite{cao2022mobiler2l} proposes a full CNN-based neural light field model and uses a SR model in its second stage, but their method is trained using an expensive distillation procedure.
RefSR-NeRF \cite{huang2023refsrnerf} proposes a specialized SR module that uses high-resolution reference image as additional input, resulting in slower training and inference times.
4K-NeRF \cite{wang20224k-nerf} synthesizes ultra high-resolution (4K) outputs using depth features as additional input and incorporates SR to achieve feasible inference times. 
Overall, these existing works 
all have high training overhead and are not meant to be optimized on lower-power consumer devices. 

In our work, we approach SR techniques for neural rendering from a different perspective. Rather than develop a complex pipeline that pushes the limit of reconstruction quality on high-end GPUs, we ask what efficiency gains, if any, can be made from a simple NeRF+SR pipeline trained on consumer-grade hardware. Our experiments show that a simple NeRF+SR pipeline can achieve comparable quality to existing complex pipelines, while being lightweight enough to train on consumer-grade hardware.

\section{Method}
\label{sec:method}

\begin{figure*}[t!]
\centering
\includegraphics[width=0.95\textwidth]{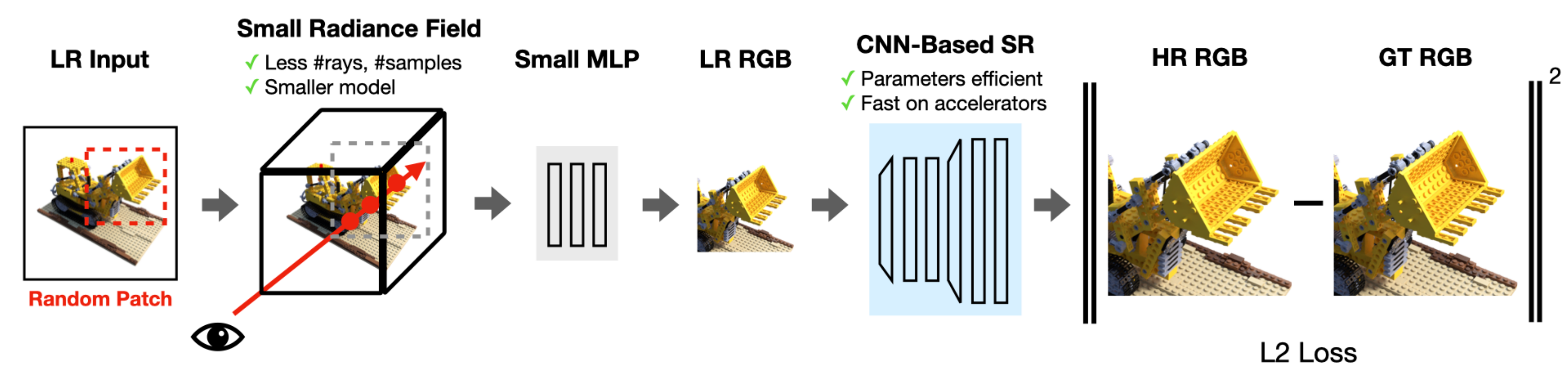} \\
\caption{Overview of \codename{}. 
The SR module in our pipeline directly takes the RGB output from NeRF model, and therefore makes our pipeline easy to implement, model agnostic and flexible to run on different devices.}
\label{fig:pipeline}
\vspace{2.5pt}
\end{figure*}


\subsection{A Simple NeRF + SR Pipeline}
\label{sec:pipeline}

As shown in Figure \ref{fig:pipeline}, our pipeline simply consists of a NeRF model concatenated with a CNN-based SR module.
Given a ray $\mathbf{r} = \mathbf{o} + t\mathbf{d}$, where $\mathbf{o}$ and $\mathbf{d}$ are respectively the ray origin and direction, NeRF reconstructs the color $\widehat{C}(\textbf{r})$ with volume rendering as follows:

\begin{equation} 
\label{eq:1}
    \widehat{C}(\mathbf{r}) = \sum_{i=1}^{N} T_{i} \cdot (1 - \exp(-\sigma_{i} \delta_{i})) \cdot c_{i},
\end{equation}
where $N$ is the number of sampling points along the ray, $\delta_{i}$ is the distance between two point sampled at $t_{i}$ and $t_{i+1}$, $T_{i} = \prod_{j=1}^{i-1} \exp(-\sigma_{j} \delta_{j})$, and $\sigma_{i}$ and $c_{i}$ are the density and color respectively of a position $\mathbf{x}$ in the 3D scene. In the original NeRF model, $\sigma_{i}$ and $c_{i}$ are computed by MLP networks $\mathcal{F}_{\sigma}$ and $\mathcal{F}_{c}$ given position and viewing direction.
%
%
%
%
%
In our pipeline, we use a hybrid NeRF model \cite{chen2022tensorf} to achieve state-of-the-art quality with improved training time and rendering speed.
To compute the density and color, we fetch radiance features from a grid-based decomposition $\mathcal{G}_{\sigma}$ and $\mathcal{G}_{c}$, and then feed sampled features to MLP $\mathcal{F}_{\sigma}$ and $\mathcal{F}_{c}$:
\begin{equation} 
\label{eq:3}
    \sigma_{i} = \mathcal{F}_{\sigma} (\mathcal{G}_{\sigma}(\mathbf{x})), 
    c_{i} = \mathcal{F}_{c}(\mathcal{G}_{c}(\mathbf{x}), \mathbf{d})
\end{equation}
Due to the more powerful discrete features, the MLPs $\mathcal{F}_{\sigma}$ and $\mathcal{F}_{c}$ in the hybrid NeRF are smaller than the ones used in vanilla NeRF. 

To train and render with the full NeRF+SR pipeline, we sample $\mathbf{r}$ from a patch of rays at low-resolution (LR) $R_{\text{LR}}^{P}$, perform volume rendering based on Equation \ref{eq:1}, and upsample the output with SR module $S$ to get the final high-resolution (HR) output $H$:
\begin{equation} 
\label{eq:4}
    \forall \mathbf{r}_{\text{LR}} \in R_{\text{LR}}^{P},   H = \mathcal{S} (\widehat{C}(\mathbf{r}_{\text{LR}};\mathcal{F};\mathcal{G}))
\end{equation}
Note that the sampling here covers a contiguous 2D patch, which differs from the stochastic sampling of rays used for training standard NeRFs. 
%
The NeRF+SR pipeline is optimized in an end-to-end manner with respect to the loss computed over the high-resolution image patch: 

\begin{equation} 
\label{eq:5}
    \mathcal{L}_{\text{MSE}} = \sum_{\mathbf{r}_{\text{HR}}, \mathbf{r}_{\text{LR}}}  \big\Vert C(\mathbf{r}_{\text{HR}}) - S(\widehat{C}(\mathbf{r}_{\text{LR}};\mathcal{F};\mathcal{G})) \big\Vert_{2}^{2}
\end{equation}

where $C$ is the ground-truth color and $\mathbf{r}_{\text{HR}}$ is the HR counterpart of $\mathbf{r}_{\text{LR}}$ in $R_{\text{HR}}^{P}$. In practice, we use bilinear interpolation to downsample $R_{\text{HR}}^{P}$ and get $R_{\text{LR}}^{P}$.

\vspace{5pt}
\noindent \textbf{Comments on Efficiency.} 
For a high-resolution NeRF model, the number of rays computed by Equation \ref{eq:1} will be $R_{\text{HR}}$, and will also require $N$ in the order of thousands to millions to reconstruct details.
In contrast, in a NeRF + SR pipeline, the NeRF only needs output a low-resolution output, which reduces the number of rays to $R_{\text{LR}}$.
$N$ can also be reduced as there are fewer details in the lower-resolution image.
In addition, we can reduce the grid and feature size of the hybrid NeRF to further improve NeRF efficiency, and still maintain the output quality at LR. 
As a result, we can greatly lower the computation overhead in Equation \ref{eq:1} and reduce the memory usage by letting NeRF output at LR.

The addition of the CNN-based SR module $S$ does not present much of a computational bottleneck.
With many years of progression on deep learning, the convolution operation is highly optimized and can efficiently run on modern commodity hardware such as GPUs \cite{cudnn}, CPUs \cite{mkldnn} and specialized accelerators \cite{ane, tensor_core}.
Furthermore, CNNs are parameter efficient by design \cite{alexnet, He2015_resnet}. 
The memory savings from down-scaling the NeRF to LR are much more than the parameters overhead induced by the SR model, which reduces overall model size. 
\subsection{Random Patch Sampling}
\label{sec:random_patch}
As discussed in Section \ref{sec:pipeline}, SR-based NeRF models need to be trained in a patch-style sampling instead of the tradition stochastic ray sampling.
Previous works handle the patch sampling by dividing the rays of an image into equal-size patches following rigid grid lines which are determined by the given grid size.
These ray patches are then shuffled and fed into the SR-based NeRF pipeline for supervision -- we call this grid-based patch sampling.
%
%
%
The problem of the grid-based patch sampling is that the sampling algorithm will cut the ray space strictly with a certain grid size.
When the model gets trained on individual patches,
there will be some variations that are never seen by the convolutional kernels as they sit between each grid boundaries.

To solve this issue, we propose \textbf{random patch sampling}.
Instead of having a rigid grid lines and restricting sampling to the grid, 
we randomly sample a region in the ray space, and use that patch to train the model.
In this way, we can still have a fixed patch size, but the content of each patch
will be different regions of the input.
After many iterations, the convolutional kernels in $S$ will cover all of the patterns appeared in the training data and lead to a better training results.

In general, CNN-based models 
are prone to over-fitting and typically require large-scale datasets \cite{deng2009imagenet} to be trained.
However, NeRF datasets usually only have tens to hundreds of training images, which is orders of magnitude smaller than a typical CNN pre-training dataset.
Existing SR-based NeRF models tackle the overfitting issue by rendering extra data from a teacher model, or guiding training with additional features from high-resolution reference image or depth maps, but these significantly increase training overhead.
Random patch sampling is a lightweight data augmentation technique that enables the convolutional kernels of the SR module to see more diversity in the training set.
This crucial augmentation allows us to achieve high-quality results without the more complex architectures or training procedures of previous works.
We provide ablation results of random patch sampling versus baselines in Section \ref{sec:eval_random_patch}.



%

\begin{table*}[tbh]
    \begin{center}
    \begin{subtable}[t]{.9\textwidth}
        \centering
        \begin{tabular}{ |m{2.75cm} ||m{1cm}|m{1cm}|m{1cm}||m{1cm}|m{1.2cm}|m{1.35cm}| }
        \hline
        \multirow{3}{*}{\textbf{Method}} & \multicolumn{6}{c|}{NeRF-Synthetic} \\
        \cline{2-7}
        & \multirow{2}{*}{\textbf{PSNR$\uparrow$}} & \multirow{2}{*}{\textbf{SSIM$\uparrow$}} & \multirow{2}{*}{\textbf{LPIPS$\downarrow$}} & \textbf{Train} & \textbf{Render} & \textbf{Model$\downarrow$} \\
        & & & & \textbf{Time$\downarrow$} & \textbf{Time(s)$\downarrow$} & \textbf{Size(MB)} \\
        \hline
        NeRF \cite{mildenhall2020nerf} & 31.01 & 0.947 & 0.081 & \textcolor{red}{$\sim$35h} & \textcolor{red}{20} & 5 \\
        TensoRF \cite{chen2022tensorf} & \textbf{33.14} & 0.963 & \textbf{0.049} & 18m & \textcolor{red}{1.4} & \textcolor{red}{71.8} \\
        \hline
        \codename{} (2$\times$) & 32.53 & 0.961 & 0.052 & 1.5h & 0.309 & 20 \\
        \codename{} (4$\times$) & 30.47 & 0.944 & 0.075 & 30m & 0.077 & 13 \\
        \codename{} (8$\times$) & 27.27 & 0.902 & 0.142 & 16m & 0.030 & 8 \\
        \hhline{|=|=|=|=||=|=|=|}
        MobileR2L \cite{cao2022mobiler2l} & 31.34 & \textbf{0.993} & 0.051 & \textcolor{red}{$>$35h} & 0.026$^{\ddagger}$ & 8.3 \\
        NeRF-SR \cite{wang2021nerf-sr} & 28.46 & 0.921 & 0.076 & \textcolor{red}{$>$35h} & \textcolor{red}{5.6} & - \\
        \hline
        \end{tabular}
        \vspace{5pt}
    \end{subtable}
    \begin{subtable}[t]{.9\textwidth}
        \centering
        \begin{tabular}{ |p{2.75cm}||p{1cm}|p{1cm}|p{1cm}||p{1cm}|p{1.2cm}|p{1.35cm}| }
            \hline
            \multirow{3}{*}{\textbf{Method}} & \multicolumn{6}{c|}{NSVF-Synthetic} \\
            \cline{2-7}
            & \multirow{2}{*}{\textbf{PSNR$\uparrow$}} & \multirow{2}{*}{\textbf{SSIM$\uparrow$}} & \multirow{2}{*}{\textbf{LPIPS$\downarrow$}} & \textbf{Train} & \textbf{Render} & \textbf{Model$\downarrow$} \\
            & & & & \textbf{Time$\downarrow$} & \textbf{Time(s)$\downarrow$} & \textbf{Size(MB)} \\
            \hline
            NeRF \cite{mildenhall2020nerf} & 30.81 & 0.952 & - & \textcolor{red}{$\sim$35h} & \textcolor{red}{$\sim$20} & $\sim$5 \\
            TensoRF \cite{chen2022tensorf} & \textbf{36.52} & 0.959 & \textbf{0.027} & 15m & \textcolor{red}{1.4} & \textcolor{red}{74} \\
            \hline
            \codename{} (2$\times$) & 35.39 & \textbf{0.979} & 0.032 & 1.5h & 0.302 & 26 \\
            \codename{} (4$\times$) & 32.04 & 0.958 & 0.059 & 30m & 0.075 & 12 \\
            \codename{} (8$\times$) & 27.93 & 0.911 & 0.119 & 16m & 0.030 & 9 \\
            \hline
        \end{tabular}
        \vspace{5pt}
    \end{subtable}
    \begin{subtable}[t]{.9\textwidth}
        \centering
        \begin{tabular}{ |p{2.75cm}||p{1cm}|p{1cm}|p{1cm}||p{1cm}|p{1.2cm}|p{1.25cm}| }
        \hline
        \multirow{3}{*}{\textbf{Method}} & \multicolumn{6}{c|}{LLFF} \\
        \cline{2-7}
        & \multirow{2}{*}{\textbf{PSNR$\uparrow$}} & \multirow{2}{*}{\textbf{SSIM$\uparrow$}} & \multirow{2}{*}{\textbf{LPIPS$\downarrow$}} & \textbf{Train} & \textbf{Render} & \textbf{Model$\downarrow$} \\
        & & & & \textbf{Time$\downarrow$} & \textbf{Time(s)$\downarrow$} & \textbf{Size(MB)} \\
        \hline
        NeRF\cite{mildenhall2020nerf} & 26.5 & 0.811 & 0.250 & \textcolor{red}{$\sim$48h} & \textcolor{red}{33} & 5 \\
        TensoRF\cite{chen2022tensorf} & 26.6 & 0.832 & 0.207 & 28m & \textcolor{red}{5.9} & \textcolor{red}{188} \\
        \hline
        \codename{} (2$\times$) & 26.20 & 0.822 & 0.241 & 2.5h & 0.786 & 26 \\
        \codename{} (4$\times$) & 25.41 & 0.784 & 0.297 & 57m & 0.165 & 15 \\
        \codename{} (8$\times$) & 21.30 & 0.584 & 0.475 & 15m & 0.040 & 8 \\
        \hhline{|=|=|=|=||=|=|=|}
        MobileR2L\cite{cao2022mobiler2l} & 26.15 & \textbf{0.966} & 0.187 & \textcolor{red}{$>$48h} & 0.018$^{\ddagger}$ & 8.3 \\
        NeRF-SR\cite{wang2021nerf-sr} & \textbf{27.26} & 0.842 & \textbf{0.103} & \textcolor{red}{$>$48h} & \textcolor{red}{39.1} & - \\ 
        RefSR-NeRF\cite{huang2023refsrnerf} & 26.23 & 0.874 & 0.243 & - & \textcolor{red}{8.5} & 38 \\
        \hline
        \end{tabular}
        \vspace{7pt}
    \end{subtable}
    \end{center}
    \vspace{-15pt}
    \caption{
        Quantitative and efficiency results on NeRF-Synthetic, NSVF-Synthetic and LLFF datasets.
        We compare the results of applying SR to the baseline NeRf model, TensoRF \cite{chen2022tensorf}, and also list vanilla NeRF as a reference.
        For NeRF-Synthetic and LLFF, we also include the results of other SR-based models from their paper.
        The results are highlighted in \textcolor{red}{red} when there is a clear disadvantage of a method.
        $\ddagger$ The rendering time is for iPhone13, while other time is on GPUs.
    }
    \label{table:quantitative_results}
\end{table*}

\section{Evaluations}


\subsection{Datasets}
We use the following three datasets for experiments.
\vspace{3pt}

\noindent \textbf{NeRF Synthetic dataset.}
The NeRF Synthetic dataset was collected along with the origin NeRF \cite{mildenhall2020nerf} paper. It contains 8 different synthetic scenes with 360$^{\circ}$  degree views produced from the Blender \cite{blender} 3D computer graphics creation framework. Each scene in this dataset contains an object with complicated details. The object is placed in the middle of the 3D space and the backgound is white. For each scene, it has 100 training images and 200 testing images of the object from different views. The resolution of the collected images are 800\(\times\)800.

\vspace{3pt}
\noindent \textbf{NSVF Synthetic dataset}
The NSVF Synthetic dastaset was released with the NSVF \cite{liu2020nsvf} paper. It has a similar setting as NeRF Synthetic dataset with gradually more complex geometry and lightening on the main object. The resolution is also at 800\(\times\)800. 

\noindent \textbf{LLFF dataset.}
LLFF dataset was collected along with the LLFF \cite{mildenhall2019llff} paper.
The scenes in this dataset were captured in the real world with foward-facing angle. 
It also has a major object placed roughly in the middle of each scene. However, different from the NeRF Synthetic dataset, the scenes in LLFF dataset have complex background depending on the captured environment.
The original resolution of the collected images are 4032\(\times\)3024, and it also provide images at 4\(\times\) and 8\(\times\)lower resolution.
Due to the practical usage, most of the NeRF works including us evaluate this dataset on 4x lower resolution at 1008\(\times\)756.
Each scene in the LLFF dataset has 20 to 40 images, and 7/8 are used for training and 1/8 are used for testing.


\subsection{Experiment Setup}
We choose TensoRF \cite{chen2022tensorf} as our NeRF backbone as it achieves state-of-the-art results on both quality and efficiency, without requiring customized CUDA kernels, and therefore aligns with the goal of this paper.
For our SR model, we chose EDSR \cite{Lim_2017_CVPR_Workshops_edsr} due to its accessible implementation and wide adoption in the computer vision community \cite{hf_edsr}.
Although we choose TensoRF and EDSR as our NeRF and SR model, both of them can be replaced with other methods, as our pipeline is model agnostic.
Since our SR module solely relies on the RGB output of the NeRF, we are able to leverage pretrained SR models.
To train our pipeline, we first warm up the TensoRF model at LR using its default hyperparameters (inherited from the official implementation) for 5K iterations.
After warming up, we plug a pretrained EDSR model with desired SR ratio into our pipeline and start training end-to-end using random patch sampling.
For the training hyperparameters, we fix the learning rate at $0.0001$, patch size at 256 and 128 for SR-2$\times$ and SR-4$\times$. We use Adam optimizer \cite{adam_opt} and train the pipeline for 150 epochs. 
For each iteration in a epoch, we only feed one patch to the pipeline.
We run our experiments on a machine that is equipped with a single NVIDIA V100 GPU with 16GB memory unless we specify the hardware platform. 


\subsection{Evaluation on Quality and Efficiency}
\noindent \textbf{Efficiency gains of utilizing SR.}
Here we evaluate the quality and efficiency gains of our simple NeRF+SR pipeline.
We list peak signal-to-noise ratio (PSNR), structural similarity (SSIM) and perceptual similarity (LPIPS) \cite{zhang2018perceptual} for quantitative quality measurements, and provide training time, rendering time and model size for efficiency evaluations.
For LPIPS, we use VGG \cite{Simonyan15_vgg16} as the backbone.
The results can be found in Table \ref{table:quantitative_results}.

As shown in Table \ref{table:quantitative_results}, comparing to the backbone TensoRF \cite{chen2022tensorf} model, applying SR can generally maintain quality at the 2x ratio and enjoy efficiency benefits in rendering time and model size.
For example, our pipeline with SR-2$\times$ only has a small 0.61dB, 1.13dB and 0.4dB PSNR drop and has near no loss on SSIM and LPIPS compared to the baseline model.
Our pipeline at SR 2$\times$ even achieves a slight improvement on SSIM for NSVF-Synthetic.
For efficiency, using 2x SR rate can improve rendering time by 4.5$\times$, 4.6$\times$ and 7.5$\times$ and reducing model size by 3.6$\times$, 2.8$\times$, and 7.2$\times$ for NeRF-Synthetic, NSVF-Synthetic and LLFF respectively.
For SR-4$\times$, we observe a more notable quality loss to the baseline compared to SR-2$\times$.
However, it can still achieve qualified results such as over 30dB PSNR on synthetic datasets and just a small 1.19dB PSNR loss on LLFF dataset.
At the mean time, with 4$\times$ SR rate, it can further improve the rendering time speedup to 18.2$\times$, 18.6$\times$ and 35.7$\times$, achieve model size reduction at 5.5$\times$, 6.2$\times$, and 12.5$\times$ for NeRF-Synthetic, NSVF-Synthetic, and LLFF.
Furthermore, the training time for SR-4$\times$ is down to 30 min for synthetic datasets and 1hr for LLFF as the model run and converge faster at this rate.
For SR-8$\times$, although the efficiency is further improved, our pipeline can not maintain a good quality at this upscaling rate.

\vspace{3pt}
\noindent \textbf{Comparing to existing SR-based NeRFs.}
We compare our simple pipeline to three existing SR-based model, which are MobileR2L \cite{cao2022mobiler2l}, NeRF-SR \cite{wang2021nerf-sr} and RefSR-NeRF \cite{huang2023refsrnerf}.
Notice that MobileR2L is light field based model and is not based on radiance field.
However, they still utilize SR to enhance rendering speed so we include it for comparison.

Comparing to them, our simple pipeline with only lightweight techniques in training achieves a very clear advantage on the training time.
At SR-2$\times$, our pipeline can be trained 23.3$\times$ and 19.2$\times$ faster than existing SR-based models on NeRF-Synthetic and LLFF, while achieving either on par or better quality.

\begin{figure*}[th!]
    \centering
    \includegraphics[width=0.975\textwidth]{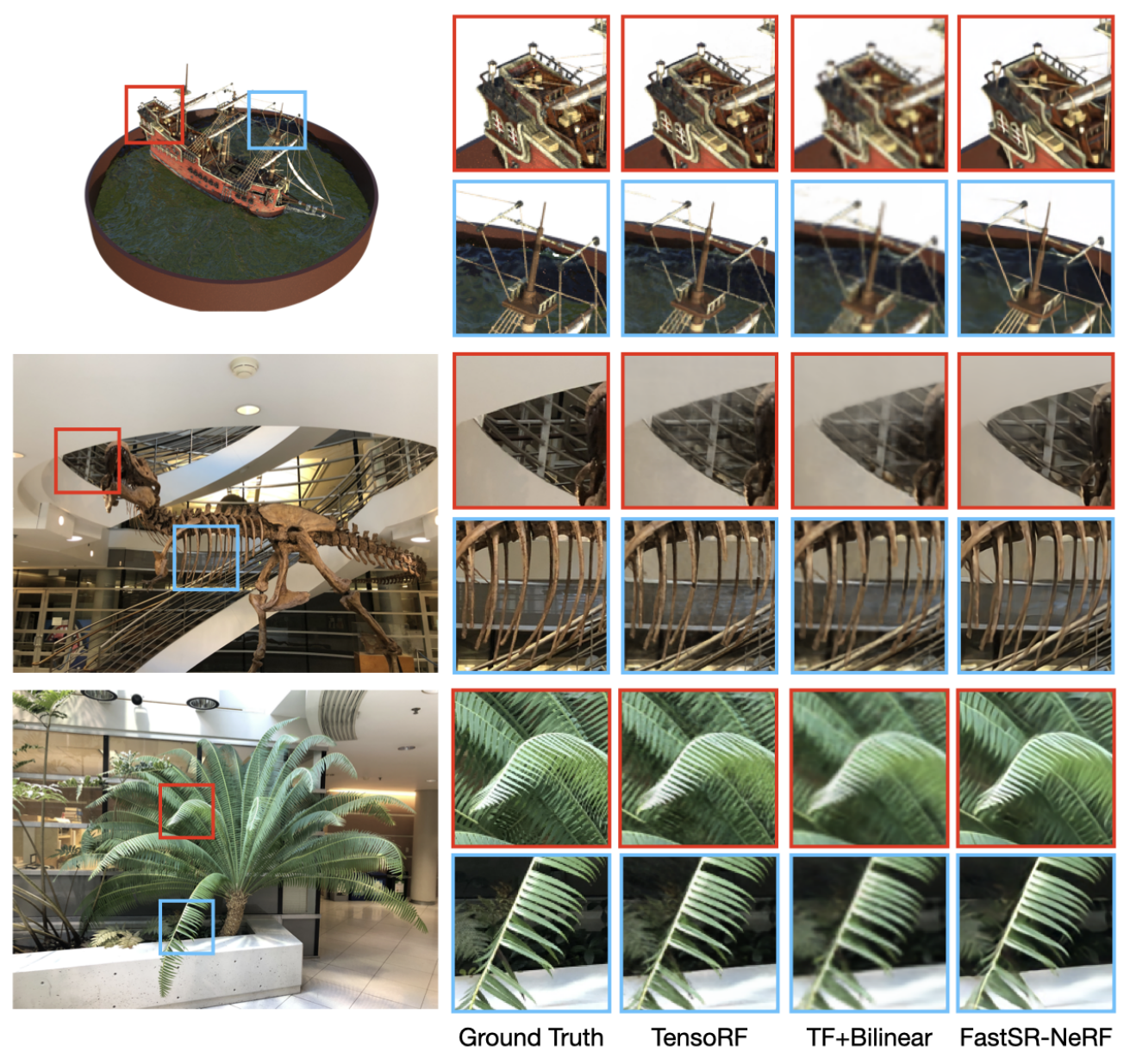}
    \vspace{-2pt}
    \caption{Qualitative results on a NeRF-Synthetic and LLFF. While TensoRF and TensoRF+Bilinear fail to recover some details (see the ribs of Trex), our pipeline successfully learn the details back with SR-2$\times$ rate. 
    }
    \vspace{-2pt} 
    \label{fig:qualitative_results}
\end{figure*}

\vspace{3pt}
\noindent \textbf{Qualitative results.} 
We show qualitative results and comparison on selected scenes from NeRF-Synthetic and LLFF datasets in Figure \ref{fig:qualitative_results}.
As Figure \ref{fig:qualitative_results} shows, with 2$\times$ upscaling rate, our pipeline can achieve on-par visual quality as the baseline TensoRF, while using bilinear interpolation at the same rate is not enough to get high fidelity results.

\begin{table}[tb]
    \centering
    \resizebox{1.\columnwidth}{!}{ 
    \renewcommand{\arraystretch}{1.1}
    \begin{tabular}{ cc|cc|cc }
        \toprule[1.5pt]
        \multirow{3}{*}{Method} & \multirow{3}{*}{PSNR} & \multicolumn{2}{c|}{M1 Pro} & \multicolumn{2}{c}{M2} \\
        & & Train & Render & Train & Render \\
        & & Time & Time & Time & Time \\
        \hline 
        TensoRF & 33.14 & 2h & 54s & 2.5h & 45.4s \\
        \hline 
        \codename{} (2$\times$) & 32.53 & 22.5h & 15.9s & 16h & 15.2s \\
        \codename{} (4$\times$) & 30.47 & 15.5h & 4.2s & 13.5h & 4s \\ 
        \hline
        MobileNeRF & 30.9 & \textcolor{red}{$>$375h}$^{\dagger}$ & 0.016s & \textcolor{red}{$>$375h}$^{\dagger}$ & 0.017s \\
        \bottomrule[1.5pt]
    \end{tabular}
    }
    \caption{PSNR and time profiling of running vanilla TensoRF, \codename{} (ours) and MobileNeRF \cite{chen2022mobilenerf} on a MacBook Pro laptop with M1 Pro chip.
    ${\dagger}$ The training time is approximated for training the vanilla NeRF on M-series CPUs, which is only the first training stage of MobileNeRF. 
    }
    \label{table:macbook_training}
    \vspace{-10pt}
\end{table}

\subsection{Training on Consumer Devices.}
We run our pipeline on a MacBook Pro with M1 Pro chip and a MacBook Air with M2 chip to evaluate efficiency on consumer platforms.  
The training time, rendering time and PSNR on NeRF-Synthetic for our pipeline at SR rate 2$\times$ and 4$\times$ are listed in Table \ref{table:macbook_training}.
We also list MobileNeRF's \cite{chen2022mobilenerf} results for comparison.

As shown in Table \ref{table:macbook_training}, using SR can improve the rendering speed by up to 3.4$\times$ and 12.8$\times$ for 2$\times$ and 4$\times$ SR rate on the consumer-grade M-series CPUs.
For MobileNeRF \cite{chen2022mobilenerf}, although it can achieve a much faster rendering time from its specialized caching mechanism, it needs more than 15 days to be trained on the same device, which is difficult to make a meaningful NeRF application that run fully on a consumer-grade platform.
In contrast, our pipeline, although can not achieve real-time rendering, it still significantly accelerates the rendering process with a reasonable training time (less than 1 day).
Note that our experiments are run on CPUs because 
the current Apple Metal Performance Shader (MPS) \cite{mps} support in PyTorch can not fully run the operators needed in NeRF and SR on the MPS device.
We expect our training and rendering speed to be faster once PyTorch has a better MPS operators support.



\subsection{The Importance of Random Patch Sampling}
\label{sec:eval_random_patch}

We evaluate the effectiveness of random patch sampling, which we discussed in Section \ref{sec:random_patch}.
To evaluate, we first establish our model pipeline as we explained in Section \ref{sec:pipeline}.
We then keep all the settings of the pipeline the same but use different patch sampling algorithm to train our model.
Notice that we fix the patch size and training for the same number of iterations, so the number of patches seen by the model is the same for both sampling methods.
For grid-based patch sampling, we randomize the order of the patches fed into the model to ensure training stability.
We report the averaged PSNR for SR ratio 2$\times$ and 4$\times$ of training our model with these two patch sampling algorithms in Table \ref{table:grid_vs_random_patch}.

As Table \ref{table:grid_vs_random_patch} shows, 
we observe that random patch sampling consistently leads to a higher PSNR than grid-based patch sampling on synthetic datasets, 
The highest improvement appear on NSVF-Synthetic when the SR rate is 4$\times$, where random patch sampling records a 1.59 PSNR increase over grid-base patch sampling.

On real-world forward facing LLFF dataset, the PSNR enhancement is less significant than the synthetic datasets.
We observe a slight PSNR decrease (0.16dB) for SR-2$\times$ but a clear PSNR increase (0.47dB) for SR-4$\times$.
We hypothesize that this is because the real-world scenes typically contains greater complexity and finer-grained detail than the synthetic scenes.
For example, in a synthetic scene, there is usually one major object and the space outside of the object is empty.
Although, the object might has some difficult and fine-grained patterns, the model can focus on learning the patterns on the object.
However, in a real world scene, although it usually still has a major object, the background is usually messy and contains a lot of small details.
Therefore, using random patches on real world scenes such as LLFF dataset can not bring a big difference in terms of the total number of patterns converged in the patches.
As a result, random patch sampling shows a greater improvement on synthetic datasets but still has the ability to enhance PSNR on real world scenes dataset when the SR rate is higher, e.g. 4$\times$.

\begin{table}[tb]
    \centering
    \resizebox{0.9\columnwidth}{!}{ 
    \begin{tabular}{ cccc }
        \toprule[1.5pt]
        \multirow{2}{*}{Dataset} & \multirow{2}{*}{Method} & \multicolumn{2}{c}{PSNR} \\
        & & SR-2$\times$ & SR-4$\times$ \\
        \midrule[0.5pt] 
        \multirow{2}{*}{NeRF-Synthetic} 
        & Grid-based & 31.84 & 29.28\\
        & Random & \textbf{32.53} & \textbf{30.47} \\
        \hline
        \multirow{2}{*}{NSVF-Synthetic} & Grid-based & 34.34 & 30.45 \\
        & Random & \textbf{35.39} & \textbf{32.04} \\
        \hline
        \multirow{2}{*}{LLFF} & Grid-based & \textbf{26.2} & 24.94 \\
        & Random & 26.04 & \textbf{25.41} \\
        \bottomrule[1.5pt]
    \end{tabular}
    }
    \caption{PSNR comparison on using \textbf{random patch sampling} versus \textbf{grid-based patch sampling}. We highlight the better results for the same SR ratio in \textbf{bold}.}
    \vspace{-10pt} 
    \label{table:grid_vs_random_patch}
\end{table}

\begin{table}[tb]
    \begin{center} 
    \resizebox{0.95\columnwidth}{!}{ 
    \begin{tabular}{ ccccc }
        \toprule[1.5pt]
        Upsample & Upsample & Train & Train & \multirow{2}{*}{PSNR} \\
        Ratio & Method & Strategy & Time(m) & \\
        \midrule[0.5pt] 
        \multirow{6}{*}{2$\times$} 
        & Bilinear & - & 11 & 29.77 \\[1ex]
        \cline{2-5} \\[-2ex]
        & \multirow{5}{*}{EDSR} 
        & Pretrained & 11 &  30.40 \\
        & & Scratch & 51 & 31.64 \\
        & & FT-GridPatch & 51 & 31.84 \\
        & & FT-RandPatch & 89 & \textbf{32.53} \\
        & & Distillation & 365 & 32.12 \\
        \hline & \\[-1.5ex]
        \multirow{6}{*}{4$\times$} 
        & Bilinear & - & 3.5 & 26.67 \\[1ex]
        \cline{2-5} \\[-2ex]
        & \multirow{5}{*}{EDSR} 
        & Pretrained & 3.5 & 27.62 \\
        & & Scratch & 19 & 29.03 \\
        & & FT-GridPatch & 19 & 29.28 \\
        & & FT-RandPatch & 30 & \textbf{30.47} \\
        & & Distillation & 166 & 29.94\\
        \bottomrule[1.5pt]
    \end{tabular}}
    \end{center}
    \vspace{-5pt}
    \caption{Training time and PSNR of different training strategies on NeRF Synthetic dataset. Pretrained EDSR model is downloaded from HuggingFace website. FT stands for finetuning the pretrained SR model. 
    RandPatch signifies random patch sampling (other methods use grid-based patch sampling if not specified).
    All experiments are trained for 150 epochs, with the exception of distillation. For distillation, we generate 1k extra training image using a pretrained TensoRF as teacher, and train for 100 epochs.
    }
    \vspace{-10pt} 
    \label{table:abalation_train_strategies}
\end{table}


\subsection{Ablation Study on Training Strategies.}
\label{sec:training_strategies}

In Section \ref{sec:eval_random_patch}, we compare the results of training our NeRF + SR pipeline with grid-based or random patch sampling.
However there are many more configurations possible.
In this section, we compare the results of 1) using bilinear interpolation as the SR method, 2) use out-of-box pretrained SR model without finetuning, 3) training the NeRF and SR model both from scratch, 4) finetuning the pipeline with pretrained SR on the NeRF dataset with grid-based patch sampling, 5) same as (4) but uses random patch sampling, 6) use the distillation method proposed by \cite{cao2022mobiler2l} and train on a larger training set augmented by a teacher NeRF at HR.

We train the pipeline on NeRF-synthetic dataset and show the comparison in Table \ref{table:abalation_train_strategies}. 
Using bilinear interpolation as SR has the shortest training time (only requires warming up the NeRF backbone) but has a significant PSNR decrease.
Directly using a pretrained EDSR model can also cut down the training time and has a better PSNR than bilinear.
However, training on NeRF dataset still help it to achieve a better accuracy.
Among those training methods, finetuning SR using random patch sampling achieves the best results while paying a little more training time (exclude distillation) due to the random sampling overhead.
For training the pipeline, although we see a promising PSNR increase with 1K extra training images generated by a teacher TensoRF, the training time becomes much longer as we need to train the NeRF model at HR first and also need to render many HR images.
Note that the PSNR of distillation might increase if we generate more training images, but the training time will be even longer.
We do not further optimize our distillation procedure as it's not the focus in this paper.
To sum up, using random patch sampling and finetuning a pretrained SR model gives us the best trade-off between time and quality under our pipeline setup.







\label{sec:exp}
\section{Conclusion}
\label{sec:conclusion}
In this work, we study the limit of SR-based NeRF model.
We propose \codename{} and find a cohesive and simple NeRF + SR pipeline can actually achieve impressive quality while also being compute and memory efficient.
The key result of this approach is that, although it's not the fastest nor the smallest model, it remains efficient for all of training time, rendering latency and model size. 
We achieve this by leveraging the lightweight technique called random patch sampling and pretrained SR model -- both of these interventions can boost our pipeline's accuracy without introducing prohibitive computational overhead.
Our pure Python-based approach (without any customized GPU kernels) allows the whole training \& inference pipeline to run on consumer-grade devices such as a laptop with a reasonable time.
We believe this work and comprehensive analysis will help the development of an end-to-end NeRF application that can purely be deployed on personal devices for improved compute efficiency and user privacy.



{\small
\bibliographystyle{ieee_fullname}
\bibliography{egbib}
}



\newpage
\begin{appendices}

\section{Additional Ablation Study on Augmentations} 

In the present study, we introduce a technique of random patch sampling designed to improve the training efficacy of the NeRF+SR pipeline. In addition to this, we extend the same framework to incorporate additional data augmentations conventionally employed in Convolutional Neural Networks (CNNs), such as random rotation or perspective transformation applied to the sampled patches. Our central hypothesis posits that these slight image transformations could potentially generate novel patterns absent from the training set but pertinent to the 3D spatial context. Through these augmentations, the SR module is hypothesized to further generalize, thereby facilitating the recovery of lost details in unobserved perspectives.


Mathematically, the NeRF+SR pipeline involves the sampling of a patch in the low-resolution (LR) ray space $R_{\text{LR}}^{P}$ and its high-resolution (HR) counterpart $R_{\text{HR}}^{P}$.
A transformation $\mathcal{A}$ is subsequently applied to these patches, yielding the transformed patches $R_{\text{LR}}^{P'}$ and $R_{\text{HR}}^{P'}$ as defined in Equation 5. These transformed patches are then integrated into Equation 3 for training the pipeline.

\begin{equation} 
\label{eq:6}
R_{\text{LR}}^{P'} = \mathcal{A}(R_{\text{LR}}^{P}), R_{\text{HR}}^{P'} = \mathcal{A}(R_{\text{HR}}^{P})
\end{equation}

To empirically validate our hypothesis, we implement two lightweight augmentations, random rotation and random horizontal flip, layered atop the random patch sampling technique. For synthetic datasets, the maximum rotation angle is set to 10 degrees, and the probability for a horizontal flip is set at 10\%. Conversely, for the real-world scenes dataset LLFF, the maximum rotation angle is limited to 5 degrees, and the random horizontal flip is omitted since it's inappropriate with the forward-facing scenes.

The empirical results, presented in Table \ref{table:abalation_aug}, reveal a marginal degradation in the output PSNR when using transformation-based augmentations as compared to utilizing random patch sampling exclusively. Consequently, random patch sampling remains as the most effective lightweight augmentation strategy for enhancing the NeRF+SR pipeline. We earmark the exploration of the effective utilization of transformation-based augmentations within the NeRF+SR pipeline for future research endeavors.

\begin{table}[h]
    \begin{center} 
    \resizebox{0.95\columnwidth}{!}{ 
    \begin{tabular}{ ccccc }
        \toprule[1.5pt]
        Dataset & Data Aug & 2x & 4x & 8x \\
        \midrule[0.5pt] 
        \multirow{3}{*}{NeRF-Synthetic} 
        & Grid-Patch & 31.84 & 29.28 & 26.02 \\
        & Rand-Patch & \textbf{32.53} & \textbf{30.47} & \textbf{27.27} \\
        & RP+RRot+Hflip & 32.22 & 30.26 & 27.26 \vspace{1pt} \\
        \hline
        \multirow{3}{*}{NSVF-Synthetic} 
        & Grid-Patch & 34.34 & 30.45 & 26.26 \\
        & Rand-Patch & \textbf{35.39} & \textbf{32.04} & \textbf{27.93} \\
        & RP+RRot+Hflip & 35.03 & 31.78 & 27.79 \vspace{1pt}\\
        \hline
        \multirow{3}{*}{LLFF} 
        & Grid-Patch & \textbf{26.2} & 24.94 & \textbf{21.68} \\
        & Rand-Patch & 26.04 & \textbf{25.41} & 21.3 \\
        & RP+RRot & 25.94 & 25.37 & 21.29 \\
        \bottomrule[1.5pt]
    \end{tabular}}
    \end{center}
    \caption{
    PSNR of using different augmentation techniques for SR rate 2x, 4x and 8x.
    The output resolution is 800x800 for NeRF-Synthetic and NSVF-Synthetic, and 1008x756 for LLFF. Grid-Patch stands for grid-based patch sampling and Rand-Patch stands for random patch sampling. RRot stands for random rotation and Hflip stands for random horizontal flip. The best results in each SR rate and dataset are highlighted in bold.
    }
    \label{table:abalation_aug}
\end{table}

\section{Additional Qualitative Results}

\begin{figure*}[th!]
    \centering
    \includegraphics[width=0.95\textwidth]{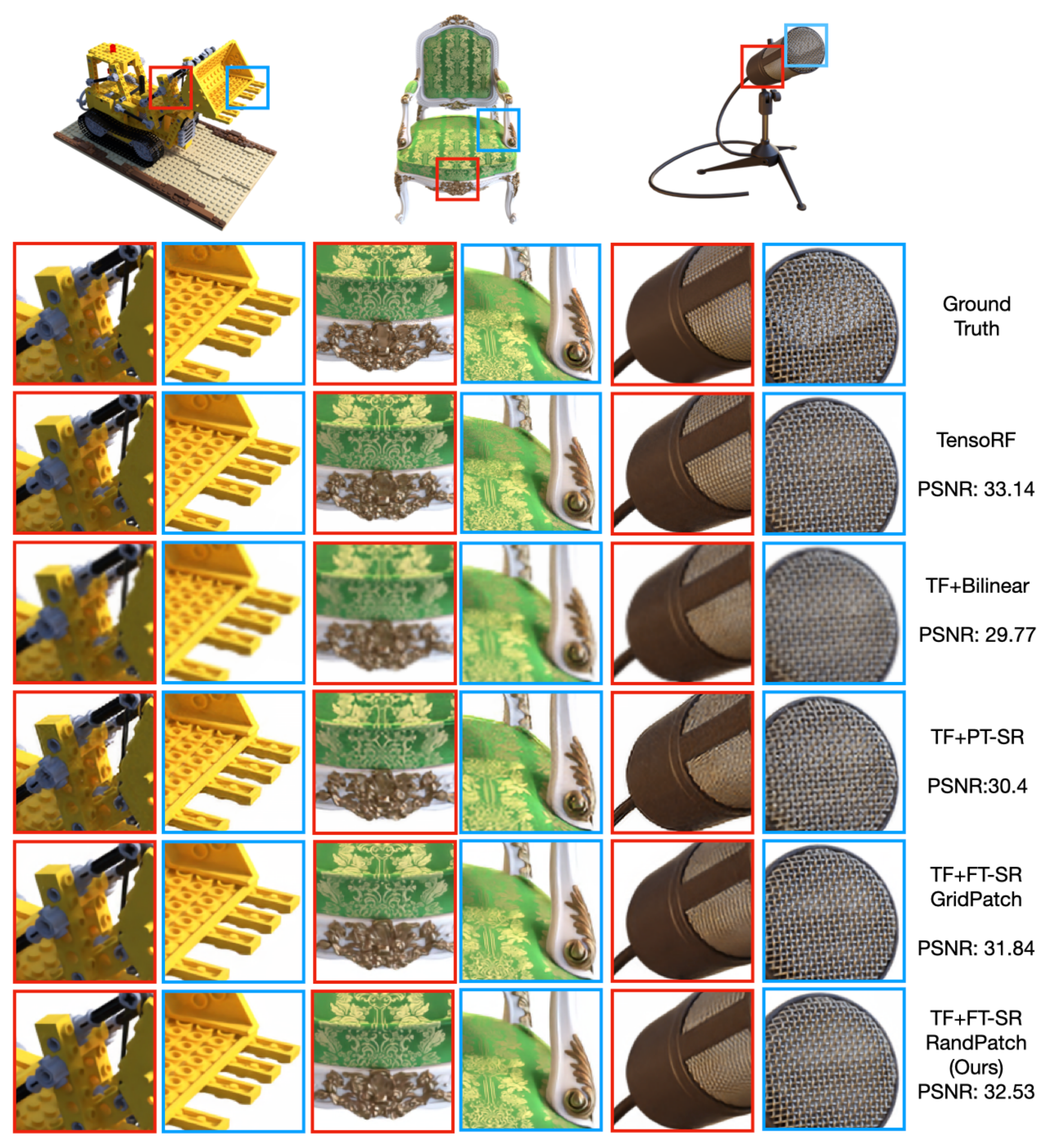}
    \vspace{-5pt}
    \caption{Qualitative results on lego, chair and mic scenes in NeRF-Synthetic. We show comparison on TensoRF at HR, and using bilinear, pretrained SR, finetuned SR with grid-based patch sampling and finetuned SR with random patch sampling to upsample output from TensoRF. The SR rate is 2$\times$, and the SR module is EDSR \cite{Lim_2017_CVPR_Workshops_edsr}.
    We show the average PSNR on NeRF-Synthetic dataset of each method.
    }
    \vspace{-8pt} 
    \label{fig:add_qualitative_results}
\end{figure*}

We show additional qualitative results in Figure \ref{fig:add_qualitative_results}. Here we compare using different SR methods and different training procedures on the SR module. 
For the SR methods, we compare using bilinear interpolation and EDSR \cite{Lim_2017_CVPR_Workshops_edsr}.
For different training procedures, we compare taking the pretrained EDSR from \cite{hf_edsr}, finetuning the EDSR model with grid-based patch sampling and finetuning the EDSR model with random patch sampling

As discerned from Figure \ref{fig:add_qualitative_results}, reliance on bilinear interpolation culminates in outputs characterized by a lack of sharpness, rendering them blurry. In contrast, utilization of a pretrained SR module yields images of greater clarity, albeit with some loss of intricate details. Subsequent finetuning facilitates the recovery of nuanced patterns, such as shadows. Remarkably, the deployment of our proposed random patch sampling methodology further enhances performance, as evidenced by improvements in the Peak Signal-to-Noise Ratio (PSNR) when compared to traditional grid-based patch sampling.

\section{Additional Comparison with more NeRF models.}

Beyond the data presented in Table 1, we extend our comparative analysis to encompass additional NeRF models specifically optimized for efficiency, incorporating both super-resolution (SR) based and non-SR-based approaches, as enumerated in Table \ref{table:additional_results}. The empirical results delineated in Table \ref{table:additional_results} affirm that our proposed pipeline not only maintains high-quality output but also excels in terms of efficiency. This efficiency is observed across multiple metrics including training duration, rendering velocity, and model compactness, all achieved without necessitating specialized CUDA support on GPUs.

\begin{table*}[h]
    \begin{center}
    \begin{subtable}[t]{1.\textwidth}
        \centering
        \begin{tabular}{ ccccccc }
        \toprule[1.5pt]
        \multirow{2}{*}{\textbf{Method}} & \multirow{2}{*}{\textbf{PSNR$\uparrow$}} & \multirow{2}{*}{\textbf{SSIM$\uparrow$}} & \multirow{2}{*}{\textbf{LPIPS$\downarrow$}} & \textbf{Train} & \textbf{Render} & \textbf{Model} \\
        & & & & \textbf{Time} & \textbf{Time(s)} & \textbf{Size(MB)} \\
        \midrule[1pt] 
        NeRF \cite{mildenhall2020nerf} & 31.01 & 0.947 & 0.081 & \textcolor{red}{$\sim$35h} 
        & \textcolor{red}{20} & 5 \\
        MipNeRF \cite{mipnerf} & 33.09 & 0.961 & 0.043 & \textcolor{red}{$\sim$35h} & - & - \\
        \hline & \\[-1.5ex]
        NSVF$^{\ddagger}$ \cite{liu2020nsvf} & 31.74 & 0.953 & 0.047 & \textcolor{red}{$>$48h} & \textcolor{red}{3} & - \\
        KiloNeRF$^{\pmb{\dagger}}$ \cite{killonerf} & 31.00 & 0.950 & \underline{0.030} & \textcolor{red}{$>$35h} & 0.026 & - \\
        SNeRG \cite{hedman2021snerg} & 30.38 & 0.950 & 0.05 & \textcolor{red}{$\sim$35h} & 0.012 & \textcolor{red}{86.8} \\
        MobileNeRF$^{\ddagger}$$^{\pmb{\dagger}}$ \cite{chen2022mobilenerf} & 30.90 & 0.947 & 0.062 & \textcolor{red}{$>$35h} & 0.0013 & \textcolor{red}{125.8} \\
        Efficient-NeRF \cite{Hu_2022_CVPR_efficient_nerf} & 31.68 &  0.954 & \textbf{0.028} & \textcolor{red}{6h} & 0.004 & \textcolor{red}{$\sim$3000} \\
        TensoRF \cite{chen2022tensorf} & \underline{33.14} & \underline{0.963} & 0.049 & 18m & \textcolor{red}{1.4} & \textcolor{red}{71.8} \\
        DVGO \cite{sun2021dvgo} & 31.95 & 0.958 & 0.053 & 14m & 0.44 & \textcolor{red}{612} \\
        FastNeRF \cite{garbin2021fastnerf} & 29.90 & 0.937 & 0.056 & - & 0.041 & \textcolor{red}{$>$7000} \\
        Plenoxel$^{\pmb{\dagger}}$ \cite{fridovich2022plenoxels} &  31.71 & 0.958 & 0.049 & 11m & 0.066 & \textcolor{red}{815} \\
        Instant-NGP$^{\pmb{\dagger}}$ \cite{muller2022instant} & \textbf{33.18} & - & - & 5m & 0.016 & 16 \\
        \hline & \\[-1.5ex]
        MobileR2L \cite{cao2022mobiler2l} & 31.34 & \textbf{0.993} & 0.051 & \textcolor{red}{$>$35h} & - & 8.3 \\
        NeRF-SR \cite{wang2021nerf-sr} & 28.46 & 0.921 & 0.076 & \textcolor{red}{$>$35h} & \textcolor{red}{5.6} & - \\
        \textbf{\codename{} ($2\times$)} & 32.53 & 0.961 & 0.052 & 1.5h & 0.309 & 20 \\
        \bottomrule[1.5pt]
        \end{tabular}
        \vspace{5pt}
        \caption{NeRF Synthetic Dataset results.}
        \vspace{7pt}
    \end{subtable}
    \begin{subtable}[t]{1.\textwidth}
        \centering
        \begin{tabular}{ ccccccc }
            \toprule[1.5pt]
            \multirow{2}{*}{\textbf{Method}} & \multirow{2}{*}{\textbf{PSNR$\uparrow$}} & \multirow{2}{*}{\textbf{SSIM$\uparrow$}} & \multirow{2}{*}{\textbf{LPIPS$\downarrow$}} & \textbf{Train} & \textbf{Render} & \textbf{Model} \\
            & & & & \textbf{Time} & \textbf{Time(s)} & \textbf{Size(MB)} \\
            \midrule[1pt] 
            NeRF \cite{mildenhall2020nerf} & 30.81 & 0.952 & - & \textcolor{red}{$\sim$35h} & \textcolor{red}{$\sim$20} & $\sim$5 \\
            \hline & \\[-1.5ex]
            NSVF$^{\ddagger}$ \cite{liu2020nsvf} & 35.13 & \textbf{0.979} & - & \textcolor{red}{$>$48h} & \textcolor{red}{$\sim$3} & - \\
            DVGO \cite{sun2021dvgo} & 35.18 & \textbf{0.979} & - & $\sim$20m & - & \textcolor{red}{$\sim$600} \\
            TensoRF \cite{chen2022tensorf} & \textbf{36.52} & 0.959 & \textbf{0.027} & 15m & \textcolor{red}{1.4} & \textcolor{red}{74} \\
            \hline & \\[-1.5ex]
            \textbf{\codename{} ($2\times$)} & \underline{35.39} & \textbf{0.979} & \underline{0.032} & 1.5h & 0.302 & 26 \\
            \bottomrule[1.5pt]
        \end{tabular}
        \vspace{5pt}
        \caption{NSVF Synthetic Dataset results.}
        \vspace{7pt}
    \end{subtable}
    \begin{subtable}[t]{1.\textwidth}
        \centering
        \begin{tabular}{ ccccccc }
            \toprule[1.5pt]
            \multirow{2}{*}{\textbf{Method}} & \multirow{2}{*}{\textbf{PSNR$\uparrow$}} & \multirow{2}{*}{\textbf{SSIM$\uparrow$}} & \multirow{2}{*}{\textbf{LPIPS$\downarrow$}} & \textbf{Train} & \textbf{Render} & \textbf{Model} \\
            & & & & \textbf{Time} & \textbf{Time(s)} & \textbf{Size(MB)} \\
            \midrule[1pt] 
            NeRF \cite{mildenhall2020nerf} & 26.5 & 0.811 & 0.250 & \textcolor{red}{$\sim$48h} & \textcolor{red}{33} & 5 \\
            \hline & \\[-1.5ex]
            SNeRG \cite{hedman2021snerg} & 25.63 & 0.818 & 0.183 & \textcolor{red}{$\sim$48h} & 0.036 & \textcolor{red}{310}\\
            NeX \cite{Wizadwongsa2021_NeX} & \underline{27.26} & \underline{0.904} & 0.178 & \textcolor{red}{20h} & 0.0033 & - \\
            Efficient-NeRF \cite{Hu_2022_CVPR_efficient_nerf} & \textbf{27.39} & \underline{0.912} & \textbf{0.082} & \textcolor{red}{4h} & 0.005 & \textcolor{red}{4300} \\
            TensoRF \cite{chen2022tensorf} & 26.6 & 0.832 & 0.207 & 28m & \textcolor{red}{5.9} & \textcolor{red}{188} \\
            \hline & \\[-1.5ex]
            MobileR2L \cite{cao2022mobiler2l} & 26.15 & \textbf{0.966} & 0.187 & \textcolor{red}{$>$48h} & - & 8.3 \\
            NeRF-SR \cite{wang2021nerf-sr} & \underline{27.26} & 0.842 & \underline{0.103} & \textcolor{red}{$>$48h} & \textcolor{red}{39.1} & - \\ 
            RefSR-NeRF \cite{huang2023refsrnerf} & 26.23 & 0.874 & 0.243 & - & \textcolor{red}{8.5} & 38 \\
            \textbf{\codename{} ($2\times$)} & 26.20 & 0.822 & 0.241 & 2.5h & 0.786 & 26 \\
            \bottomrule[1.5pt]
        \end{tabular}
        \vspace{5pt}
        \caption{LLFF Dataset results.}
        \vspace{7pt}
    \end{subtable}
    \end{center}
    \vspace{-15pt}
    \caption{
        Quality and efficiency results on NeRF-Synthetic, NSVF-Synthetic, and LLFF datasets. The tables are organized into three sections: implicit MLP-based NeRFs, efficient fully-explicit or hybrid NeRFs, and SR-based NeRFs, including our approach. Top performance in each quantitative metric is marked in bold, and the second best is underlined. Clear efficiency disadvantages are highlighted in red. Our tests run on a NVIDIA V100 GPU, while other results are their GPU results cited from respective papers when available.
        $\ddagger$ notes such method requires 8 high-end GPUs to train. $\pmb{\dagger}$ notes the method relies on customized CUDA kernels.
        Our method produces \textbf{excellent quantitative results that is on par or only slightly less than the state-of-the-art} cross all the benchmarks. Our method further achieves \textbf{great efficiency results across training time, rendering speed and model size} without the need of customized CUDA kernels support, which is favorable for inexpensive consumer-grade devices.
    }
    \label{table:additional_results}
\end{table*}

\end{appendices}

\end{document}